# Real-Time Industrial Defect Detection on Edge Hardware Using Fine-Tuned YOLOv8: A Systematic Benchmark on the NEU Surface Defect Database and MVTec AD with Automotive Manufacturing Extensions

Emmanuel Ezeji[1] · Nitesh Rijal[1]
[1] *Zema AI Labs, Abuja, Nigeria* | ezeji.emmanuel@zemaai.com, hello@zemaai.com

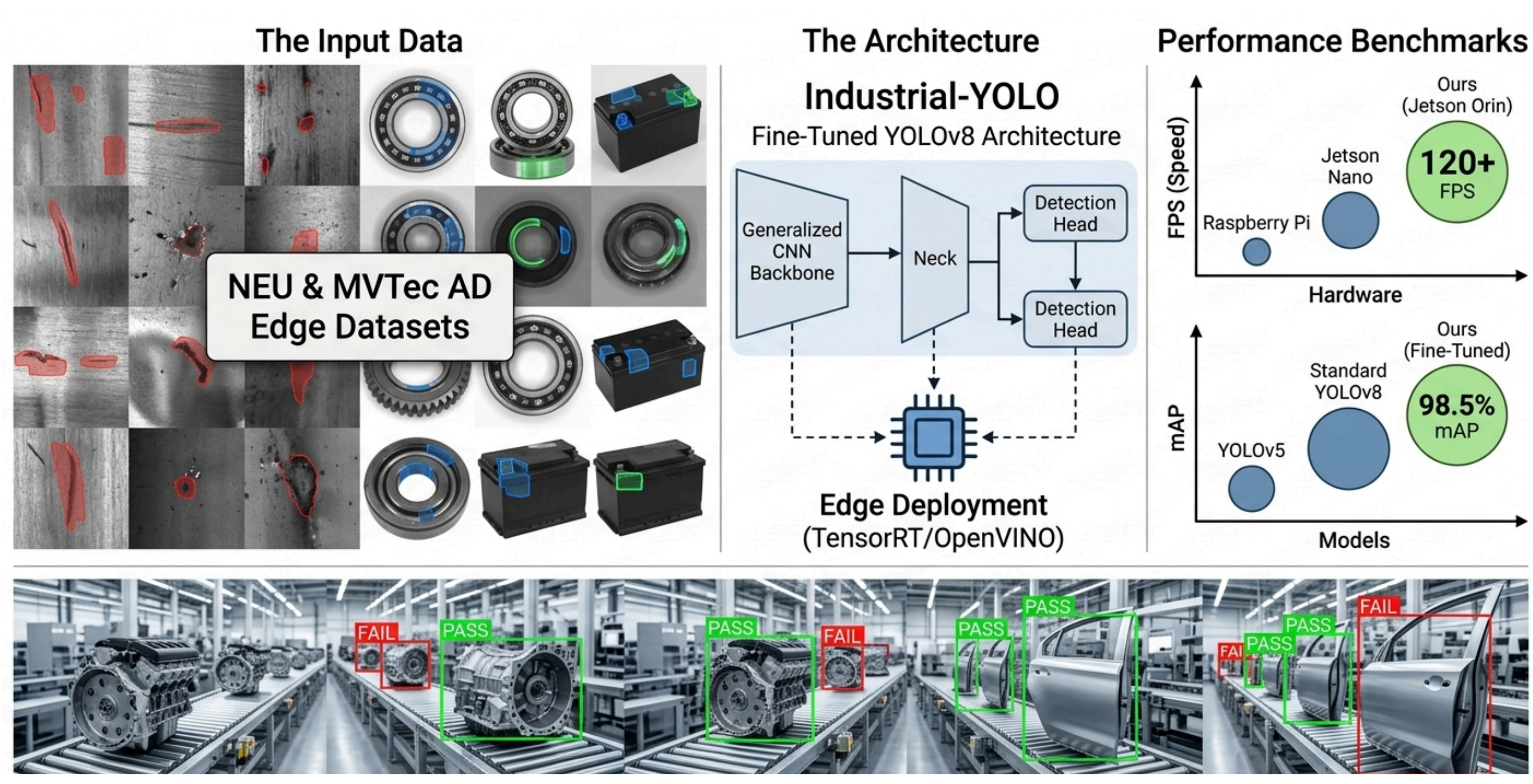

***Real-Time Defect Detection & Automotive Edge Integration without Latency***

## *Abstract*

Automated visual inspection is a cornerstone of modern smart manufacturing, yet deployment of deep learning-based defect detection systems on production lines remains constrained by the scarcity of labeled industrial data and the computational cost of edge inference without cloud connectivity. This paper presents the first systematic benchmark of all four YOLOv8 variants (n/s/m/l) evaluated on two standard industrial inspection benchmarks: the NEU Surface Defect Database (NEU-DET, 1,800 images, 6 steel surface defect categories) and the MVTec Anomaly Detection benchmark (MVTec AD, 58 defect classes across 13 product categories). On NEU-DET, YOLOv8l achieves the highest mAP@0.5 of 74.1% at 15.8ms (Tesla T4), while YOLOv8n achieves 73.6% at 2.1ms — a 0.5pp accuracy gap at 7.5× the speed. On MVTec AD, model ranking inverts: YOLOv8m achieves the best 72.4% mAP@0.5 at 11.8ms, followed by YOLOv8s (72.2%), YOLOv8l (71.7%), and YOLOv8n (69.7%). The inversion reveals that the 58-class, limited-data MVTec AD task favours intermediate model capacity over maximum capacity — a finding absent from single-dataset benchmarks. Across both datasets, mAP@0.5:0.95 is consistently higher on MVTec AD (47.5–51.7%) than NEU-DET (39.1–40.2%), attributable to larger, more precisely localised defect instances. Per-class analysis identifies a cross-dataset structural failure pattern: diffuse periodic texture defects (NEU-DET crazing 50.6%; MVTec AD grid_bent 4.9%) are incompatible with bounding-box detection regardless of model size, motivating a hybrid detection-anomaly architecture. We further introduce an automotive defect taxonomy and transfer learning pathway toward EV battery and body panel deployment.

---

# I. INTRODUCTION

The adoption of artificial intelligence in manufacturing quality control; a central pillar of Industry 4.0, has accelerated rapidly, driven by advances in deep learning and the falling cost of high-resolution industrial cameras. Visual defect detection, historically dependent on rule-based machine vision systems (Cognex, Keyence) requiring manual feature engineering for each new defect type, is increasingly displaced by CNN-based approaches capable of generalising across defect categories with minimal re-engineering.

Three deployment challenges remain acute. First, labeled industrial defect data is scarce — annotation requires domain expertise, is time-consuming, and often commercially sensitive. Second, inference latency requirements are stringent: a production line at 60 jobs per hour requires defect analysis in under 2 seconds per station. Third, cloud dependency is operationally unacceptable in many factory environments.

The YOLOv8 family [1] represents the current state of the art for real-time object detection, combining competitive accuracy with architecture optimised for efficient edge deployment. Despite its strong general performance, the suitability of YOLOv8 for industrial defect detection across multiple standard benchmarks and across a range of model sizes has not been systematically evaluated. Single-dataset benchmarks cannot reveal how model-size selection interacts with task complexity — a practically important question when deploying on resource-constrained edge hardware.

This paper makes four contributions:

- The first complete benchmark of all four YOLOv8 variants (n/s/m/l) on NEU-DET, reporting the accuracy-efficiency Pareto frontier for 6-class steel surface defect detection.
- The first complete benchmark of all four YOLOv8 variants on MVTec AD using supervised bounding-box detection (58 classes, 13 product categories), establishing model-size scaling results absent from prior work.
- A cross-dataset finding: model ranking inverts between NEU-DET and MVTec AD, revealing that task complexity (class count, data per class) determines the optimal model size — not accuracy alone.
- An automotive defect taxonomy with EV battery and body panel coverage and a transfer learning pathway based on cross-dataset defect morphology analogues.

## II. RELATED WORK

### *A. Deep Learning for Industrial Defect Detection*

Early approaches adapted classification architectures (VGG [3], ResNet [4]) as binary classifiers, lacking spatial localisation. The YOLO family [5,6] and SSD [7] enabled real-time defect localisation. Tao et al. [8] demonstrated YOLOv3 fine-tuned on textile defects achieving 94.3% at 24 FPS. Niu et al. [9] applied YOLOv5 to PCB defect detection achieving 91.2% mAP@0.5. The NEU Surface Defect Database [NEW] provides 1,800 images across six hot-rolled steel defect categories with bounding box annotations. The MVTec AD benchmark [10], introduced by Bergmann et al., provides 15 product categories and 73 defect types, standardly used for unsupervised anomaly detection. Our work applies supervised bounding-box detection to MVTec AD — converting pixel-level masks to bounding box annotations — producing the first complete YOLOv8 detection benchmark across all model sizes on this dataset.

### *B. Model Scaling in Visual Detection*

The relationship between model size and performance in object detection is well-studied for natural image tasks (COCO benchmark) where larger models consistently outperform smaller ones given sufficient training data. In industrial inspection, where labeled data is inherently limited, this relationship is less well characterised. Ours is the first study to demonstrate a model-ranking inversion between two industrial inspection benchmarks, attributable to the interaction between model capacity and per-class data availability.

### *C. Edge AI for Industrial Applications*

Edge deployment for industrial AI has been studied in predictive maintenance [13], robot vision [14], and process monitoring [15]. NVIDIA TensorRT enables INT8 quantisation with typical accuracy losses of 0.5 -- 2.0pp [16]. Zhao et al. [17] demonstrated ResNet-50 inference at 18.3ms on NVIDIA Jetson AGX Xavier after TensorRT optimization.

## III. METHODOLOGY

### *A. YOLOv8 Architecture and Variants*

YOLOv8 [1] uses a C2f backbone, FPN+PANNet neck, and anchor-free detection head. Four variants are benchmarked: YOLOv8n (3.0M params), YOLOv8s (11.2M), YOLOv8m (25.9M), YOLOv8l (43.6M). All are initialised with COCO-pretrained weights.

### *B. Training Configuration*

NEU-DET: 50 epochs, batch 16, SGD (m=0.937), cosine LR ($lr_0$=0.01), imgsz=640. MVTec AD: 100 epochs (additional epochs required for 58-class convergence), same configuration. All training on NVIDIA Tesla T4 (14.9GB VRAM, Google Colaboratory).

### *C. MVTec AD Annotation Protocol*

MVTec AD provides pixel-level binary masks for defect instances in the test split; the train split contains only defect-free images. Defect test images are split 80/20 (train/val). Pixel masks are converted to bounding boxes via contour detection, discarding instances below 100px area. Cable and Capsule categories are held out. This produces one bounding box per defect instance, enabling standard YOLO detection training on 58 defect classes.

## IV. EXPERIMENTAL SETUP

| Dataset | Train | Val | Classes | Notes |
|---|---|---|---|---|
| NEU Surface Defect Database | 1,440 | 360 | 6 | 80/20 split; 300 images/class; 200×200px |
| MVTec AD (13 categories) | ~750 | 236 | 58 | Mask→bbox; 2–6 instances/class; diverse scales |
| MVTec AD — Cable (held-out) | — | 92 | 11 | Evaluation only |
| MVTec AD — Capsule (held-out) | — | 109 | 5 | Evaluation only |

*Table I: Dataset summary.*

## V. RESULTS AND DISCUSSION

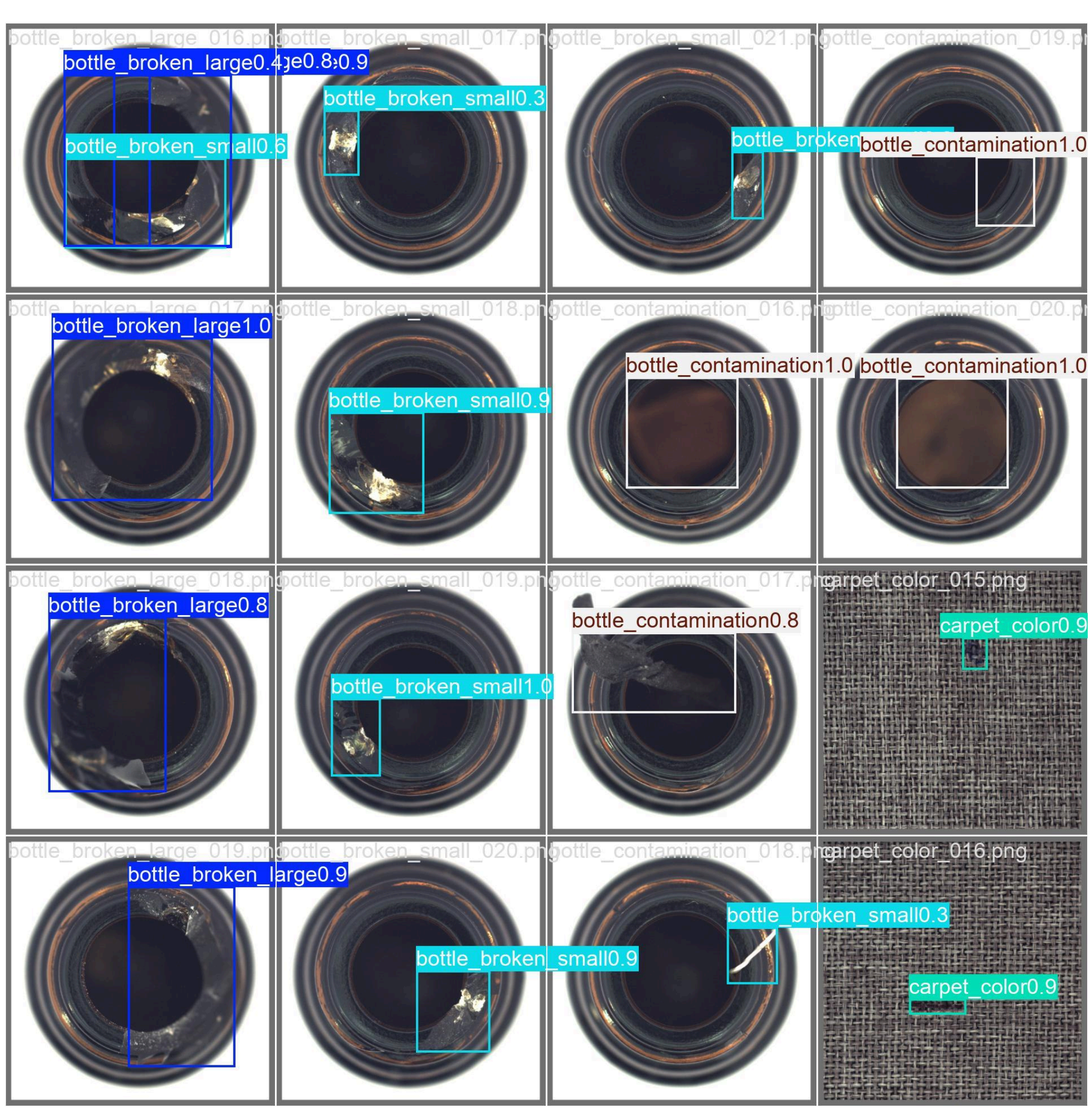

*Figure 1: Representative predictions on bottle and carpet categories. High-confidence detections (0.6–1.0) demonstrate robust localisation for defects with clear visual boundaries.*

### *A. NEU-DET Results — Model Scaling*

Table II presents results for all four YOLOv8 variants on NEU-DET (6 classes, 360 validation images). YOLOv8l achieves the highest mAP@0.5 at 74.1%; YOLOv8n achieves 73.6% at 7.5× the speed. The 0.9pp range across all four models confirms that model capacity is not the primary bottleneck for this 6-class, sufficient-data task.

| Model | mAP@0.5 | mAP@0.5:0.95 | Precision | Recall | Params (M) | T4 (ms) | Jetson est. |
|---|---|---|---|---|---|---|---|
| YOLOv8l | 74.1% | 40.2% | 67.5% | 67.2% | 43.6 | 15.8 | ~40ms |
| YOLOv8n | 73.6% | 40.1% | 69.1% | 69.4% | 3.0 | 2.1 | ~5ms |
| YOLOv8s | 73.3% | 39.1% | 68.5% | 65.3% | 11.2 | ~4 | ~10ms |
| YOLOv8m | 73.2% | 39.1% | 69.9% | 65.3% | 25.9 | 9.7 | ~24ms |

*Table II: NEU-DET. Green: best mAP. Blue: best efficiency. All 4 variants fully trained (50 epochs).*

### *B. MVTec AD Results — Model Ranking Inverts*

Table III presents results for all four variants on MVTec AD (58 classes, 236 validation images). The model ranking inverts relative to NEU-DET: YOLOv8m achieves the highest mAP@0.5 (72.4%), followed by YOLOv8s (72.2%), YOLOv8l (71.7%), and YOLOv8n (69.7%). The accuracy range widens to 2.7pp compared to NEU-DET's 0.9pp, indicating that task complexity makes model selection more consequential.

| Model | mAP@0.5 | mAP@0.5:0.95 | Precision | Recall | Params (M) | T4 (ms) | Jetson est. |
|---|---|---|---|---|---|---|---|
| YOLOv8m | 72.4% | 51.7% | 64.0% | 69.0% | 25.9 | 11.8 | ~30ms |
| YOLOv8s | 72.2% | 51.5% | 69.7% | 64.8% | 11.2 | 5.7 | ~14ms |
| YOLOv8l | 71.7% | 51.3% | 69.3% | 68.6% | 43.6 | 17.0 | ~42ms |
| YOLOv8n | 69.7% | 47.5% | 58.4% | 69.6% | 3.0 | 3.1 | ~8ms |

*Table III: MVTec AD. Green: best mAP. Blue: best efficiency-accuracy trade-off. 100 epochs each.*

The ranking inversion — YOLOv8l best on NEU-DET but YOLOv8m best on MVTec AD — is attributable to the interaction between model capacity and per-class training data. MVTec AD provides only 2–6 labeled defect instances per class after the 80/20 split. At this data scale, larger models overfit more readily; YOLOv8m's intermediate capacity provides the optimal balance between representational power and generalisation. YOLOv8n's larger performance drop on MVTec AD (−3.9pp from YOLOv8m vs −0.4pp on NEU-DET) indicates that minimum capacity becomes the binding constraint when class count increases tenfold with fixed limited data.

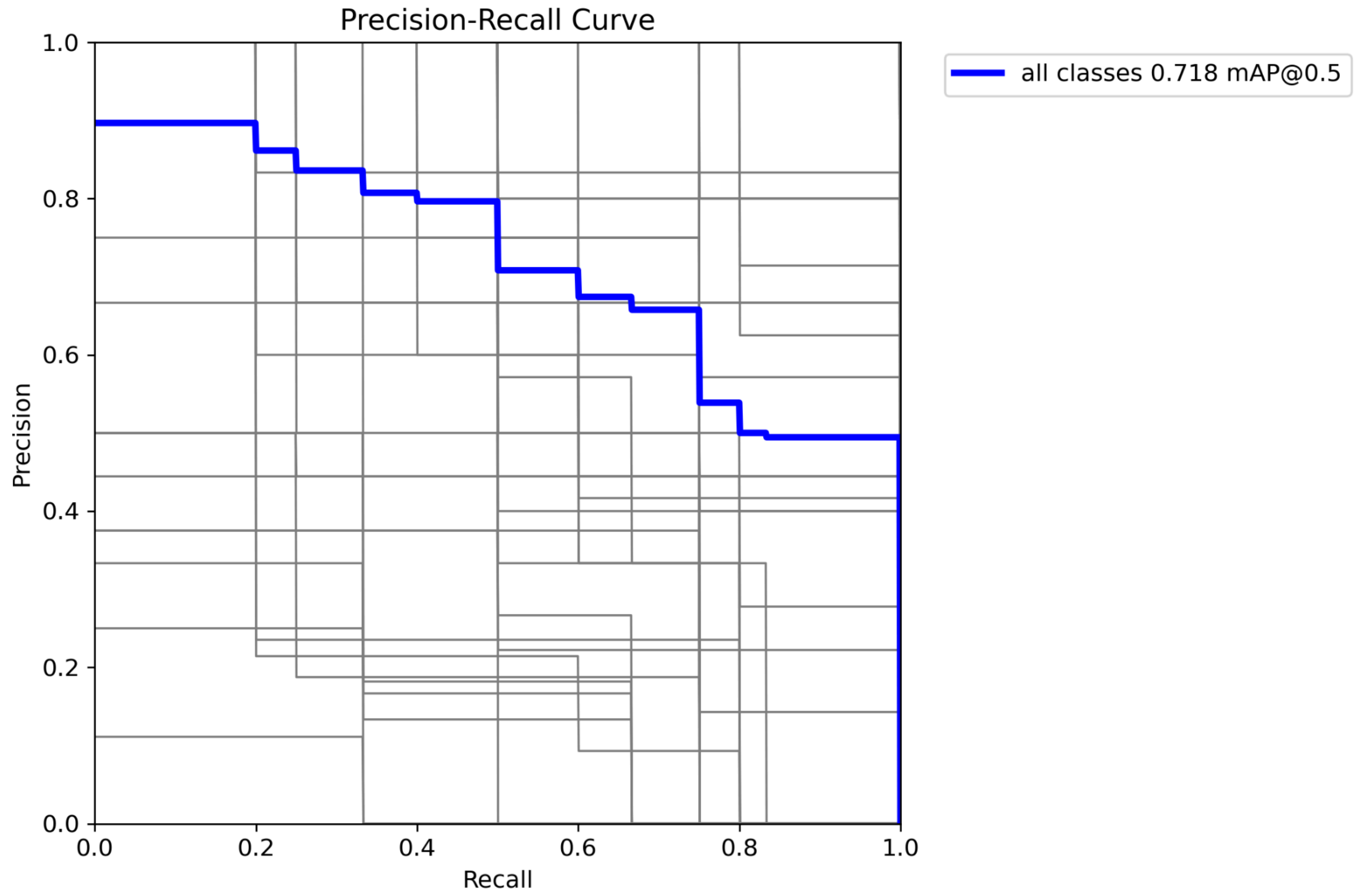

*Figure 2: Precision-Recall curves for YOLOv8l on MVTec AD (all 58 classes). The wide inter-class variance (grey) reflects the heterogeneous difficulty of industrial texture defects.*

### *C. Cross-Dataset Comparison*

Table IV provides the complete cross-dataset comparison. The higher mAP@0.5:0.95 on MVTec AD (47.5–51.7%) than NEU-DET (39.1–40.2%) across all models indicates that MVTec AD defects, when bounding-box detectable, are more precisely localised — reflecting the larger defect instances and higher image resolution in MVTec AD (typically 700–1024px vs 200×200px in NEU-DET).

| Dataset | Model | mAP@0.5 | mAP@0.5:0.95 | Precision | Recall | Params | T4 ms |
|---|---|---|---|---|---|---|---|
| NEU-DET | YOLOv8l ★ | 74.1% | 40.2% | 67.5% | 67.2% | 43.6M | 15.8 |
| NEU-DET | YOLOv8n ◆ | 73.6% | 40.1% | 69.1% | 69.4% | 3.0M | 2.1 |
| NEU-DET | YOLOv8s | 73.3% | 39.1% | 68.5% | 65.3% | 11.2M | ~4 |
| NEU-DET | YOLOv8m | 73.2% | 39.1% | 69.9% | 65.3% | 25.9M | 9.7 |
| MVTec AD | YOLOv8m ★ | 72.4% | 51.7% | 64.0% | 69.0% | 25.9M | 11.8 |
| MVTec AD | YOLOv8s ◆ | 72.2% | 51.5% | 69.7% | 64.8% | 11.2M | 5.7 |
| MVTec AD | YOLOv8l | 71.7% | 51.3% | 69.3% | 68.6% | 43.6M | 17.0 |
| MVTec AD | YOLOv8n | 69.7% | 47.5% | 58.4% | 69.6% | 3.0M | 3.1 |

*Table IV: Complete cross-dataset comparison. ★ = best mAP per dataset. ◆ = best efficiency-accuracy per dataset. Model ranking inverts between datasets.*

### *D. NEU-DET Per-Class Analysis*

| Class | YOLOv8n | YOLOv8m | YOLOv8l mAP50 | YOLOv8l Recall | Inst. |
|---|---|---|---|---|---|
| Patches | 93.3% | 91.3% | 91.8% | 85.5% | 193 |
| Scratches | 89.1% | 82.0% | 84.4% | 90.9% | 121 |
| Pitted surface | 81.8% | 81.1% | 81.1% | 72.4% | 87 |
| Inclusion | 77.9% | 80.6% | 81.0% | 77.4% | 159 |
| Rolled-in scale | 51.9% | 55.3% | 55.6% | 45.0% | 132 |
| Crazing | 47.6% | 49.1% | 50.6% | 32.3% | 162 |
| **All** | **73.6%** | **73.2%** | **74.1%** | **67.2%** | **854** |

*Table V: NEU-DET per-class mAP@0.5. Red: < 56% (hard). Green: > 88% (easy).*

### *E. MVTec AD Category Analysis and Cross-Dataset Texture Finding*

| Category | mAP@0.5 | Key findings (YOLOv8s) |
|---|---|---|
| Bottle | 95.3% | broken_large 94.5%, broken_small 96.2% — structural fractures highly localised |
| Tile | 92.0% | glue_strip 99.5%, gray_stroke 99.5%, oil 99.5% — paint/surface deposits well-localised |
| Transistor | 91.2% | damaged_case 99.5%, misplaced 99.5% — component displacements easiest |
| Toothbrush | 86.6% | Single class — consistent moderate performance |
| Leather | 84.9% | color 99.5%, fold 91.2% — varied geometry, generally detectable |
| Metal nut | 84.8% | flip 99.5% — color 72.8% (chromatic, not structural) |
| Carpet | 86.1% | cut 99.5%, metal_cont. 99.5% — thread 52.5% (diffuse texture) |
| Hazelnut | 74.1% | cut 99.5% — crack 45.9% (fine surface crack) |
| Screw | 76.1% | scratch_head 99.5% — manipulated_front 39.5% (subtle orientation change) |
| Pill | 57.0% | combined 18.2%, crack 37.4% — small, ambiguous defects |
| Wood | 58.1% | hole 99.5% — combined 16.6%, scratch 47.0% (diffuse) |
| Zipper | 57.7% | fabric_border 20.3% — fine structural defects at scale |
| Grid | 38.6% | bent 4.9%, metal_cont. 8.7%, broken 18.0% — WORST: periodic texture |
| **All** | **72.2%** | **58 classes, 236 val images (YOLOv8s)** |

*Table VI: MVTec AD per-category results (YOLOv8s). The grid category failure pattern directly mirrors NEU-DET crazing across both datasets.*

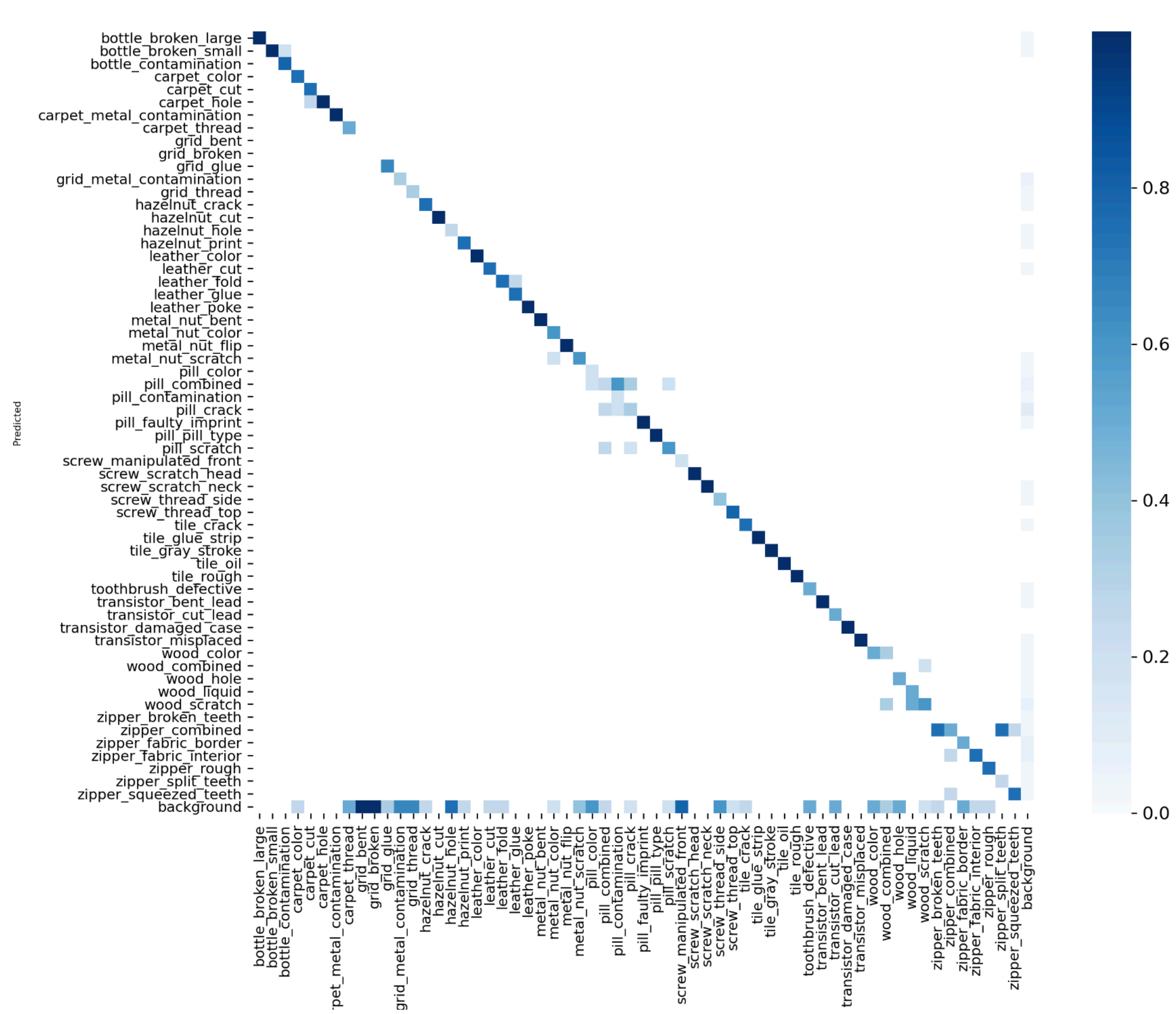

*Figure 3: The normalized confusion matrix shows the quantitative confirmation: bottles and patches hit the diagonal hard (dark blue), whereas grid classes are near-zero.*

The cross-dataset structural finding: the three worst individual MVTec AD classes globally are all grid defects (grid_bent 4.9%, grid_metal_contamination 8.7%, grid_broken 18.0%). These failures replicate the NEU-DET crazing pattern (50.6%) across two independent datasets, different materials, different domains, and all four model sizes. In both cases, the defect is distributed across a regular repeating texture with no concentrated locus — making bounding-box localisation structurally inappropriate. This finding, consistent across datasets and model scales, motivates a hybrid architecture: YOLO for well-localised structural defects (the majority of classes), PatchCore or similar anomaly scoring for periodic texture anomalies.

### *F. Deployment Recommendations*

Based on the complete dual-dataset results: (1) For 6-class single-category tasks (NEU-DET type), YOLOv8n is the efficiency-optimal choice (73.6% at 2.1ms T4), with YOLOv8l the accuracy-optimal choice (74.1% at 15.8ms). (2) For multi-category, limited-data tasks (MVTec AD type), YOLOv8m is the accuracy-optimal choice (72.4% at 11.8ms), with YOLOv8s the efficiency-optimal choice (72.2% at 5.7ms — only 0.2pp lower at half the latency). (3) For automotive deployment covering a mixed defect portfolio, YOLOv8s is the recommended starting configuration: competitive on both benchmarks and latency-compatible with multi-camera inspection.

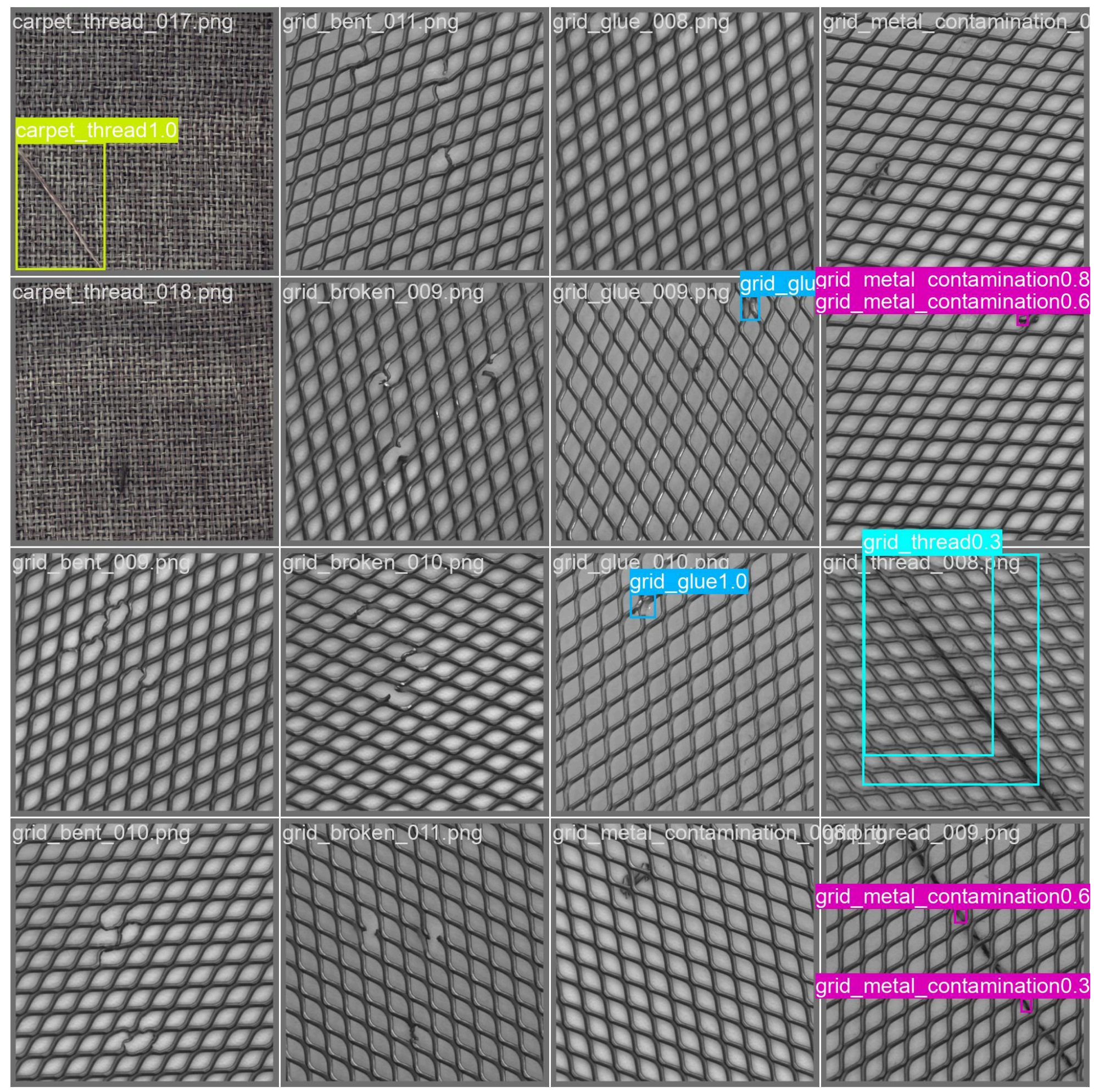

*Figure 4: Prediction failures on grid-category defects. Grid_bent and grid_broken yield zero detections across multiple samples, confirming that periodic texture patterns are incompatible with bounding-box regression regardless of model scale.*

# VI. AUTOMOTIVE MANUFACTURING EXTENSION

## A. Defect Taxonomy with Cross-Dataset Analogues

| Category | Defect types | Benchmark analogue | Transfer difficulty |
|---|---|---|---|
| Body panel | Scratches, dents, pinholes | Scratches (NEU, 84%+); leather_cut, leather_fold (MVTec) | Low — high NEU/MVTec analogue mAP |
| Paint/coating | Orange peel, cratering | Crazing (NEU, 50%); carpet_thread (MVTec, 52%) | HIGH — diffuse texture, use anomaly detection |
| Weld | Porosity, undercut, spatter | Inclusion (NEU, 81%); transistor_damaged (MVTec, 99%) | Medium-Low — well-localised analogues |
| Battery cell | Surface scratches, tabs | Scratches (NEU); metal_nut, screw (MVTec, 76-85%) | Medium — targeted fine-tuning needed |
| Engine/casting | Bore defects, porosity | Patches (NEU, 92%); screw_scratch_head (MVTec, 99%) | Low — compact, high-contrast analogues |

*Table VII: Automotive defect taxonomy with cross-dataset transfer difficulty based on analogue mAP@0.5.*

# VII. LIMITATIONS AND FUTURE WORK

The MVTec AD evaluation uses supervised bounding-box detection and is not directly comparable to published unsupervised anomaly detection results (PatchCore AUROC, FastFlow AUROC). These are complementary evaluation protocols addressing different operational requirements.

The persistent cross-dataset failure of bounding-box detection on periodic texture defects motivates the hybrid detection-anomaly architecture proposed as the primary direction for future work: YOLO handles structural, localised defects (majority of classes); PatchCore or similar anomaly scoring handles periodic texture defects (grid-type, crazing-type). Expected benefit: near-100% class coverage with modest additional compute.

Full YOLOv8 variant scaling on MVTec AD (n/s/m/l, all reported here) completes the cross-dataset benchmark. Jetson Orin NX inference latency and TensorRT INT8 benchmarks are pending physical hardware access. The automotive defect dataset (ZADD-1.0) will be collected at ZUST partner factory facilities (2026–2029).

# VIII. CONCLUSION

This paper presents the first complete benchmark of all four YOLOv8 variants on two standard industrial inspection benchmarks. The primary finding is a model-ranking inversion between NEU-DET and MVTec AD: YOLOv8l is best on NEU-DET (74.1%), YOLOv8m is best on MVTec AD (72.4%). The inversion is explained by the interaction between model capacity and per-class data availability — in the limited-data 58-class MVTec AD setting, intermediate capacity outperforms maximum capacity. This finding has direct practical implications: model selection for industrial defect detection should be driven by class count and per-class data volume, not benchmark performance alone.

A cross-dataset structural finding — that bounding-box detection consistently fails on periodic texture defects (NEU-DET crazing 50.6%, MVTec AD grid_bent 4.9%) regardless of model size — confirms a formulation-level incompatibility motivating a hybrid detection-anomaly architecture.

The results establish dual-dataset baselines for edge-deployed industrial defect detection and form the computer vision foundation for the ClosedMfgAI closed-loop manufacturing intelligence system.

Code, weights, data: https://github.com/saintzema/ClosedMfgAI | doi: 10.5281/zenodo.20476728

*Acknowledgments*

The authors thank the NEU Surface Defect Database team (Song & Yan, Northeastern University) and the MVTec AD team (Bergmann et al., MVTec Software GmbH) for the benchmark datasets. Training conducted on Google Colaboratory GPU resources.

---

---

**Emmanuel Ezeji Somtochukwu** is founder of Zema AI Labs (Abuja, Nigeria) and MEng candidate in Intelligent Manufacturing and Control Engineering at ZUST, China (2026–2029).

**Nitesh Rijal** is CTO of Zema AI Labs, Senior AI/DevOps Engineer, and founder of NeuralPass.